%% file: code_gen.tex
\documentclass[11pt,a4paper]{article}
\usepackage[hyperref]{acl2017}
\usepackage{times}
\usepackage{latexsym}
\input{settings.tex}

\aclfinalcopy 
\def\aclpaperid{086} 

\expandafter\def\expandafter\normalsize\expandafter{%
    \normalsize
    \setlength\abovedisplayskip{4pt}
    \setlength\belowdisplayskip{4pt}
    \setlength\abovedisplayshortskip{4pt}
    \setlength\belowdisplayshortskip{4pt}
}

\title{A Syntactic Neural Model for General-Purpose Code Generation}

\author{Pengcheng Yin \\
  Language Technologies Institute \\
  Carnegie Mellon University \\
  {\tt pcyin@cs.cmu.edu} \\\And
  Graham Neubig \\
  Language Technologies Institute \\
  Carnegie Mellon University \\
  {\tt gneubig@cs.cmu.edu} \\}

\date{}

\begin{document}
\maketitle
\begin{abstract}
  We consider the problem of parsing natural language descriptions into source code written in a general-purpose programming language like Python.
  Existing data-driven methods treat this problem as a language generation task without considering the underlying syntax of the target programming language.
  Informed by previous work in semantic parsing, in this paper we propose a novel neural architecture powered by a grammar model to explicitly capture the target syntax as prior knowledge.
  Experiments find this an effective way to scale up to generation of complex programs from natural language descriptions, achieving state-of-the-art results that well outperform previous code generation and semantic parsing approaches.
\end{abstract}

\section{Introduction}
\label{sec:intro}

Every programmer has experienced the situation where they know what they want to do, but do not have the ability to turn it into a concrete implementation.
For example, a Python programmer may want to ``{\it sort my\_list in descending order},'' but not be able to come up with the proper syntax {\tt sorted(my\_list, reverse=True)} to realize his intention.
To resolve this impasse, it is common for programmers to search the web in natural language (NL), find an answer, and modify it into the desired form \cite{DBLP:conf/chi/BrandtGLDK09,DBLP:conf/chi/BrandtDWK10}.
However, this is time-consuming, and thus the software engineering literature is ripe with methods to directly generate code from NL descriptions, mostly with hand-engineered methods highly tailored to specific programming languages~\cite{DBLP:journals/tse/Balzer85,DBLP:journals/ase/LittleM09,DBLP:conf/icse/GveroK15}.

In parallel, the NLP community has developed methods for data-driven semantic parsing, which attempt to map NL to structured logical forms executable by computers.
These logical forms can be general-purpose meaning representations \cite{DBLP:journals/coling/ClarkC07,DBLP:conf/acllaw/BanarescuBCGGHK13}, formalisms for querying knowledge bases \cite{DBLP:dblp_conf/ecml/TangM01,ZettlemoyerC05,berant2013freebase} and instructions for robots or personal assistants \cite{artzi-zettlemoyer:2013:TACL,DBLP:conf/acl/QuirkMG15}, among others.
While these methods have the advantage of being learnable from data, compared to the programming languages (PLs) in use by programmers, the {\it domain-specific} languages targeted by these works have a schema and syntax that is relatively simple.

Recently, \newcite{DBLP:conf/acl/LingBGHKWS16} have proposed a data-driven code generation method for high-level, {\it general-purpose} PLs like Python and Java.
This work treats code generation as a sequence-to-sequence modeling problem, and introduce methods to generate words from character-level models, and copy variable names from input descriptions.
However, unlike most work in semantic parsing, it does not consider the fact that code has to be well-defined programs in the target syntax.


In this work, we propose a data-driven syntax-based neural network model tailored for generation of general-purpose PLs like Python.
In order to capture the strong underlying syntax of the
PL, we define a model that transduces an NL statement into an Abstract Syntax Tree (AST; Fig.~\ref{fig:decoder_example}(a), \autoref{sec:problem}) for the target PL.
ASTs can be deterministically generated for all well-formed programs using standard parsers provided by the PL, and thus give us a way to obtain syntax information with minimal engineering.
Once we generate an AST, we can use deterministic generation tools to convert the AST into surface code.
We hypothesize that such a structured approach has two benefits.



\begin{table*}[t]
  \centering
  \small
  \renewcommand{\tabcolsep}{4pt}
  \begin{tabular}{l|l|p{2.8in}}
  \toprule
  \textbf{Production Rule} & \textbf{Role} & \textbf{Explanation} \\
  \hline
  Call $\mapsto$ expr[{\it func}] expr*[{\it args}] keyword*[{\it keywords}] & Function Call & 
    $\triangleright$ {\it func}: the function to be invoked
    $\triangleright$ {\it args}: arguments list
    $\triangleright$ {\it keywords}: keyword arguments list
    \\
  If $\mapsto$ expr[{\it test}] stmt*[{\it body}] stmt*[{\it orelse}] & If Statement &  
    $\triangleright$ {\it test}: condition expression
    $\triangleright$ {\it body}: statements inside the If clause
    $\triangleright$ {\it orelse}: elif or else statements
    \\
  For $\mapsto$ expr[{\it target}] expr*[{\it iter}] stmt*[{\it body}] & For Loop & \multirow{2}{*}{\pbox{2.8in}{ 
    $\triangleright$ {\it target}: iteration variable
    $\triangleright$ {\it iter}: enumerable to iterate over
    $\triangleright$ {\it body}: loop body
    $\triangleright$ {\it orelse}: else statements
  }}  \\
  \hspace{8.5mm} stmt*[{\it orelse}] & & \\
  FunctionDef $\mapsto$ identifier[{\it name}] arguments*[{\it args}] & Function Def. & \multirow{2}{*}{\pbox{2.8in}{ 
    $\triangleright$ {\it name}: function name
    $\triangleright$ {\it args}: function arguments \\
    $\triangleright$ {\it body}: function body
  }}  \\
  \hphantom{xxxxxxxxxxxxx} stmt*[{\it body}] & & \\
  \bottomrule
  \end{tabular}
  \caption{Example production rules for common Python statements~\cite{pythonast}}
  \label{tab:python_grammar}
  \vspace{-3mm}
\end{table*}

First, we hypothesize that structure can be used to constrain our search space, ensuring generation of well-formed code.
To this end, we propose a syntax-driven neural code generation model.
The backbone of our approach is a {\it grammar model} (\autoref{sec:grammar_model}) which formalizes the generation story of a derivation AST into sequential application of {\it actions} that either apply production rules (\autoref{sec:apply_rule}), or emit terminal tokens (\autoref{sec:gen_token}).
The underlying syntax of the PL is therefore encoded in the grammar model {\it a priori} as the set of possible actions.
Our approach frees the model from recovering the underlying grammar from limited training data, and instead enables the system to focus on learning the compositionality among existing grammar rules.
\newcite{DBLP:conf/acl/XiaoDG16} have noted that this imposition of structure on neural models is useful for semantic parsing, and we expect this to be even more important for general-purpose PLs where the syntax trees are larger and more complex.

Second, we hypothesize that structural information helps to model information flow within the neural network, which naturally reflects the recursive structure of PLs.
To test this,
we extend a standard recurrent neural network (RNN) decoder to allow for additional neural connections which reflect the recursive structure of an AST (\autoref{sec:decoder}).
As an example, when expanding the node $\star$ in Fig.~\ref{fig:decoder_example}(a), we make use of the information from both its parent and left sibling (the dashed rectangle).
This enables us to locally pass information of relevant code segments via neural network connections, resulting in more confident predictions.

Experiments (\autoref{sec:experiment}) on two Python code generation tasks show 11.7\% and 9.3\% absolute improvements in accuracy against the state-of-the-art system~\cite{DBLP:conf/acl/LingBGHKWS16}. Our model also gives competitive performance on a standard semantic parsing benchmark.  

\vspace{-2mm}
\section{The Code Generation Problem}
\label{sec:problem}
\vspace{-2mm}

Given an NL description $x$, our task is to generate the code snippet $c$ in a modern PL based on the intent of $x$.
We attack this problem by first generating the underlying AST. 
We define a probabilistic grammar model of generating an AST $y$ given $x$: $p(y|x)$.
The best-possible AST $\hat{y}$ is then given by
\begin{equation}
  \hat{y} = \argmax_{y} p(y|x).
\label{eq:code_gen_main}
\end{equation}
$\hat{y}$ is then deterministically converted to the corresponding surface code $c$.\footnote{We use {\tt astor} library to convert ASTs into Python code.}
While this paper uses examples from Python code, our method is PL-agnostic.

Before detailing our approach, we first present a brief introduction of the Python AST and its underlying grammar.
The Python abstract grammar contains a set of production rules, and an AST is generated by applying several production rules composed of a head node and multiple child nodes.
For instance, the first rule in Tab.~\ref{tab:python_grammar} is used to generate the function call {\tt sorted($\cdot$)} in Fig.~\ref{fig:decoder_example}(a).
It consists of a head node of type {\tt Call}, and three child nodes of type {\tt expr}, {\tt expr*} and {\tt keyword*}, respectively.
Labels of each node are noted within brackets.
In an AST, non-terminal nodes sketch the general structure of the target code, while terminal nodes can be categorized into two types: {\it operation terminals} and {\it variable terminals}.
Operation terminals correspond to basic arithmetic operations like {\tt AddOp}.
Variable terminal nodes store values for variables and constants of built-in data types\footnote{{\tt bool}, {\tt float}, {\tt int}, {\tt str}.}.
For instance, all terminal nodes in Fig.~\ref{fig:decoder_example}(a) are variable terminal nodes.

\vspace{-2mm}
\section{Grammar Model}
\label{sec:grammar_model}
\vspace{-2mm}

\begin{figure*}[tb]
  \centering
  \includegraphics[width=\textwidth]{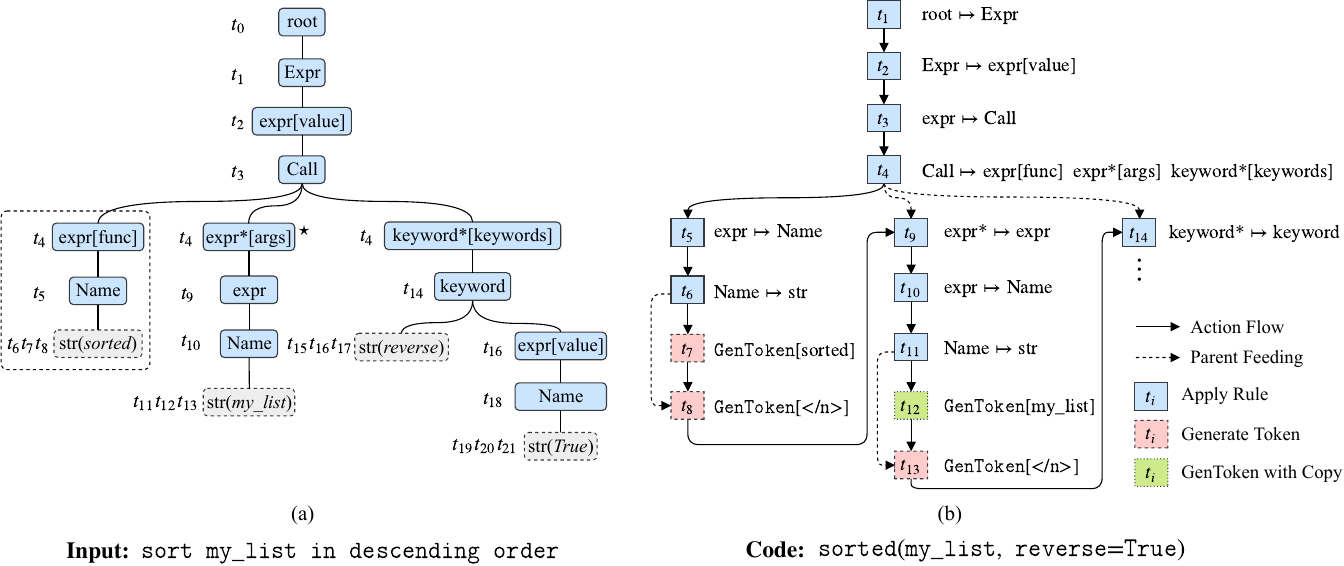}
  \vspace{-6mm}
  \caption{(a) the Abstract Syntax Tree (AST) for the given example code. 
  Dashed nodes denote terminals.
  Nodes are labeled with time steps during which they are generated.
  (b) the action sequence (up to $t_{14}$) used to generate the AST in (a) }
  \label{fig:decoder_example}
  \vspace{-3.3mm}
\end{figure*}

Before detailing our neural code generation method, we first introduce the grammar model at its core.
Our probabilistic grammar model defines the generative story of a derivation AST.
We factorize the generation process of an AST into sequential application of {\it actions} of two types: 
\begin{itemize}
  \item $\textsc{ApplyRule}[r]$ applies a production rule $r$ to the current derivation tree;
  \item $\textsc{GenToken}[v]$ populates a variable terminal node by appending a terminal token $v$.
\end{itemize}
Fig.~\ref{fig:decoder_example}(b) shows the generation process of the target AST in Fig.~\ref{fig:decoder_example}(a).
Each node in Fig.~\ref{fig:decoder_example}(b) indicates an action.
Action nodes are connected by solid arrows which depict the chronological order of the action flow. The generation proceeds in depth-first, left-to-right order (dotted arrows represent parent feeding, explained in \autoref{sec:decoder_state}).

Formally, under our grammar model, the probability of generating an AST $y$ is factorized as:
\begin{equation}
  p(y|x) = \prod_{t=1}^{T} p(a_t|x, a_{<t}),
  \label{eq:ast_prob}
\end{equation}
where $a_t$ is the action taken at time step $t$, and $a_{<t}$ is the sequence of actions before $t$.
We will explain how to compute Eq.~\eqref{eq:ast_prob} in \autoref{sec:model}.
Put simply, the generation process begins from a {\tt root} node at $t_0$, and proceeds by 
the model choosing {\sc ApplyRule} actions to generate the overall program structure from a closed set of grammar rules, then at leaves of the tree corresponding to variable terminals, the model switches to {\sc GenToken} actions to generate variables or constants from the open set.
We describe this process in detail below.

\vspace{-1.5mm}
\subsection{\textbf{\textsc{ApplyRule}} Actions}
\label{sec:apply_rule}
\vspace{-0.5mm}

{\sc ApplyRule} actions generate program structure, expanding the current node (the \textit{frontier node} at time step $t$: $n_{f_t}$) in a depth-first, left-to-right traversal of the tree. 
Given a fixed set of production rules, {\sc ApplyRule} chooses a rule $r$ from the subset that has a head matching the type of $n_{f_t}$, and uses $r$ to expand $n_{f_t}$ by appending all child nodes specified by the selected production. 
As an example, in Fig.~\ref{fig:decoder_example}(b), the rule {\tt Call} $\mapsto$ {\tt expr}$\ldots$ expands the frontier node {\tt Call} at time step $t_4$, and its three child nodes {\tt expr}, {\tt expr*} and {\tt keyword*} are added to the derivation. 

{\sc ApplyRule} actions grow the derivation AST by appending nodes. When a variable terminal node (e.g., {\tt str}) is added to the derivation and becomes the frontier node, the grammar model then switches to {\sc GenToken} actions to populate the variable terminal with tokens.

\textbf{Unary Closure} Sometimes, generating an AST requires applying a chain of unary productions.
For instance, it takes three time steps ($t_9 - t_{11}$) to generate the sub-structure {\tt expr*} $\mapsto$ {\tt expr} $\mapsto$ {\tt Name} $\mapsto$ {\tt str} in Fig.~\ref{fig:decoder_example}(a).
This can be effectively reduced to one step of {\sc ApplyRule} action by taking the closure of the chain of unary productions and merging them into a single rule: {\tt expr*} $\mapsto^*$ {\tt str}.
Unary closures reduce the number of actions needed, but would potentially increase the size of the grammar. 
In our experiments we tested our model both with and without unary closures (\autoref{sec:experiment}).


\vspace{-1.4mm}
\subsection{\textbf{\textsc{GenToken}} Actions}
\label{sec:gen_token}
\vspace{-0.5mm}


Once we reach a frontier node $n_{f_t}$ that corresponds to a variable type (e.g., {\tt str}), {\sc GenToken} actions are used to fill this node with values.
For general-purpose PLs like Python, variables and constants have values with one or multiple tokens.
For instance, a node that stores the name of a function (e.g., {\tt sorted}) has a single token, while a node that denotes a string constant (e.g., {\tt a=`hello world'}) could have multiple tokens.
Our model copes with both scenarios by firing {\sc GenToken} actions at one or more time steps.
At each time step, {\sc GenToken} appends one terminal token to the current frontier variable node. 
A special {\tt </n>} token is used to ``close'' the node.
The grammar model then proceeds to the new frontier node. 

Terminal tokens can be generated from a pre-defined vocabulary, or be directly copied from the input NL.
This is motivated by the observation that the input description often contains out-of-vocabulary (OOV) variable names or literal values that are directly used in the target code.
For instance, in our running example the variable name {\tt my\_list} can be directly copied from the the input at $t_{12}$.
We give implementation details in \autoref{sec:action_prob}.



\vspace{-1mm}
\section{Estimating Action Probabilities}
\label{sec:model}
\vspace{-1mm}

We estimate action probabilities in Eq.~\eqref{eq:ast_prob} using attentional neural encoder-decoder models with an information flow structured by the syntax trees.
%
%
\vspace{-1mm}
\subsection{Encoder}
\label{sec:encoder}
For an NL description $x$ consisting of $n$ words $\{ w_i\}_{i=1}^n$,
the encoder computes a context sensitive embedding $\mathbf{h_i}$ for each $w_i$ using a bidirectional Long Short-Term Memory (LSTM) network~\cite{DBLP:journals/neco/HochreiterS97}, similar to the setting in~\cite{DBLP:journals/corr/BahdanauCB14}.
See supplementary materials for detailed equations.

\vspace{-1mm}
\subsection{Decoder}
\label{sec:decoder}

The decoder uses an RNN to model the sequential generation process of an AST defined as Eq.~\eqref{eq:ast_prob}.
Each action step in the grammar model naturally grounds to a time step in the decoder RNN.
Therefore, the action sequence in Fig.~\ref{fig:decoder_example}(b) can be interpreted as unrolling RNN time steps, with solid arrows indicating RNN connections.
The RNN maintains an internal state to track the generation process (\autoref{sec:decoder_state}), which will then be used to compute action probabilities $p(a_t|x, a_{<t})$ (\autoref{sec:action_prob}).

\subsubsection{Tracking Generation States}
\label{sec:decoder_state}

\begin{figure}[t]
  \centering
  \includegraphics[width=\columnwidth]{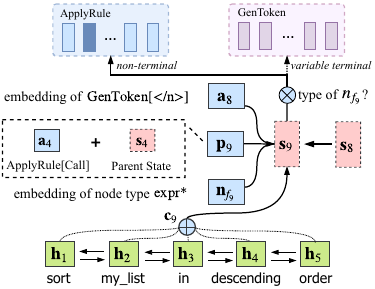}
  \caption{Illustration of a decoder time step ($t=9$)}
  \label{fig:decoder_time_step}
  \vspace{-3mm}
\end{figure}


Our implementation of the decoder resembles a vanilla LSTM, with additional neural connections 
(parent feeding, Fig.~\ref{fig:decoder_example}(b))
to reflect the topological structure of an AST.
The decoder's internal hidden state at time step $t$, $\mathbf{s}_t$, is given by:
\begin{equation}
  \mathbf{s}_t = f_{\rm LSTM} ([\mathbf{a}_{t-1}: \mathbf{c}_t: \mathbf{p}_t: \mathbf{n}_{f_t}], \mathbf{s}_{t-1}),
  \label{eq:decoder_lstm}
\end{equation}
where $f_{\rm LSTM}(\cdot)$ is the LSTM update function. 
$[:]$ denotes vector concatenation. $\mathbf{s}_t$ will then be used to compute action probabilities $p(a_t|x, a_{<t})$ in Eq.~\eqref{eq:ast_prob}.
Here, $\mathbf{a}_{t-1}$ is the embedding of the previous action. $\mathbf{c}_t$ is a context vector retrieved from input encodings $\{ \mathbf{h}_i \}$ via soft attention.
$\mathbf{p}_t$ is a vector that encodes the information of the parent action.
$\mathbf{n}_{f_t}$ denotes the node type embedding of the current frontier node $n_{f_t}$\footnote{We maintain an embedding for each node type.}. 
Intuitively, feeding the decoder the information of $n_{f_t}$ helps the model to keep track of the frontier node to expand.

\noindent {\bf Action Embedding} $\mathbf{a}_t$~
We maintain two action embedding matrices, $\mathbf{W}_R$ and $\mathbf{W}_G$.
Each row in $\mathbf{W}_R$ ($\mathbf{W}_G$) corresponds to an embedding vector for an action $\textsc{ApplyRule}[r]$ ($\textsc{GenToken}[v]$). 


\noindent {\bf Context Vector} $\mathbf{c}_t$~ The decoder RNN uses soft attention to retrieve a context vector $\mathbf{c}_t$ from the input encodings $\{ \mathbf{h}_i \}$ pertain to the prediction of the current action. 
We follow \newcite{DBLP:journals/corr/BahdanauCB14} and use a Deep Neural Network (DNN) with a single hidden layer to compute attention weights.

\noindent {\bf Parent Feeding} $\mathbf{p}_t$~
Our decoder RNN 
uses additional neural connections to directly pass information from parent actions. 
For instance, when computing $\mathbf{s}_9$, the information from its parent action step $t_4$ will be used.
Formally, we define the {\it parent action step} $p_t$ as the time step at which the frontier node $n_{f_t}$ is generated.
As an example, for $t_9$, its parent action step $p_9$ is $t_4$, since $n_{f_9}$ is the node $\star$, which is generated at $t_4$ by the $\textsc{ApplyRule}[${\tt Call}$\mapsto$$\ldots]$ action.

We model parent information $\mathbf{p}_t$ from two sources:
(1) the hidden state of parent action $\mathbf{s}_{p_t}$,
and (2) the embedding of parent action $\mathbf{a}_{p_t}$.
$\mathbf{p}_t$ is the concatenation.
The parent feeding schema enables the model to utilize the information of parent code segments to make more confident predictions.
Similar approaches of injecting parent information were also explored in the \sq/ model in~\newcite{DBLP:conf/acl/DongL16}\footnote{\sq/ generates tree-structured outputs by conditioning on the hidden states of parent non-terminals, while our parent feeding uses the states of parent actions.}.



\subsubsection{Calculating Action Probabilities}
\label{sec:action_prob}

In this section we explain how action probabilities $p(a_t|x, a_{<t})$ are computed based on $\mathbf{s}_t$.

\noindent \textbf{\textsc{ApplyRule}} 
The probability of applying rule $r$ as the current action $a_t$ is given by a softmax\footnote{We do not show bias terms for all softmax equations.}:
\begin{multline}
 p(a_t = \textsc{ApplyRule}[r]|x, a_{<t}) = \\ {\rm softmax}(\mathbf{W}_R \cdot g(\mathbf{s}_t))^\intercal \cdot \mathbf{e}(r)
 \label{eq:applyrule}
\end{multline}
where $g(\cdot)$ is a non-linearity $\tanh (\mathbf{W} \cdot \mathbf{s}_t + \mathbf{b})$, and
$\mathbf{e}(r)$ the one-hot vector for rule $r$.



\noindent \textbf{\textsc{GenToken}}
As in \autoref{sec:gen_token}, a token $v$ can be generated from a predefined vocabulary or copied from the input, defined as the marginal probability:
\begin{multline*}
   p(a_t = \textsc{GenToken}[v]|x, a_{<t}) =  \\
   p(\text{gen}|x, a_{<t})p(v|\text{gen}, x, a_{<t}) \\ + p(\text{copy}| x, a_{<t})p(v|\text{copy}, x, a_{<t}).
   \label{eq:gen_token_prob}
\end{multline*}
The selection probabilities $p(\text{gen}|\cdot)$ and $p(\text{copy}|\cdot)$ are given by ${\rm softmax}(\mathbf{W}_S \cdot \mathbf{s}_t)$.
The probability of generating $v$ from the vocabulary, $p(v|\text{gen}, x, a_{<t})$, is defined similarly as Eq.~\eqref{eq:applyrule}, except that we use the $\textsc{GenToken}$ embedding matrix $\mathbf{W}_G$, and we concatenate the context vector $\mathbf{c}_t$ with $\mathbf{s}_t$ as input.
To model the copy probability, 
we follow recent advances in modeling copying mechanism in neural networks~\cite{DBLP:conf/acl/GuLLL16,DBLP:conf/acl/JiaL16,DBLP:conf/acl/LingBGHKWS16}, and use a pointer network~\cite{DBLP:conf/nips/VinyalsFJ15} to compute the probability of copying the $i$-th word from the input by attending to input representations $\{ \mathbf{h}_i \}$:
\begin{equation*}
  p(w_i|\text{copy}, x, a_{<t}) = \frac{\exp ( \omega (\mathbf{h}_i, \mathbf{s}_t, \mathbf{c}_t) )}{\sum_{i'=1}^n \exp (\omega (\mathbf{h}_{i'}, \mathbf{s}_t, \mathbf{c}_t) )},
\end{equation*}
where $\omega(\cdot)$ is a DNN with a single hidden layer.
Specifically, if $w_i$ is an OOV word (e.g., {\tt my\_list}, which is represented by a special {\tt <unk>} token in encoding), we directly copy the actual word $w_i$ to the derivation.






\vspace{-1mm}
\subsection{Training and Inference}
\vspace{-1mm}
\label{sec:train_inference}

Given a dataset of pairs of NL descriptions $x_i$ and code snippets $c_i$, we parse $c_i$ into its AST $y_i$ and decompose $y_i$ into a sequence of oracle actions under the grammar model. The model is then optimized by maximizing the log-likelihood of the oracle action sequence.
At inference time, we use beam search to approximate the best AST $\hat{y}$ in Eq.~\eqref{eq:code_gen_main}. See supplementary materials for the pseudo-code of the inference algorithm.

\vspace{-2mm}
\section{Experimental Evaluation}
\label{sec:experiment}
\vspace{-1.5mm}


\subsection{Datasets and Metrics}

\begin{table}[]
\centering
\resizebox{0.94 \columnwidth}{!}{
\begin{tabular}{llll}
\toprule \
\textbf{Dataset} & {\sc HS} & {\sc Django} & {\sc Ifttt} \\ \midrule
Train & 533 & 16,000 & 77,495 \\
Development & 66 & 1,000 & 5,171 \\
Test & 66 & 1,805 & 758 \\ \midrule
Avg. tokens in description & 39.1 & 14.3 & 7.4 \\
Avg. characters in code & 360.3 & 41.1 & 62.2 \\ 
Avg. size of AST (\# nodes) & 136.6 & 17.2 & 7.0 \\ \midrule \midrule
\multicolumn{4}{c}{Statistics of Grammar} \\
{\bf w/o unary closure} & & & \\
\# productions & 100 & 222 & 1009 \\ 
\# node types & 61 & 96 & 828 \\
terminal vocabulary size & 1361 & 6733 & 0 \\  
Avg. \# actions per example & 173.4 & 20.3 & 5.0 \\ \midrule
{\bf w/ unary closure} & & & \\
\# productions & 100 & 237 & -- \\ 
\# node types & 57 & 92 & -- \\
Avg. \# actions per example & 141.7 & 16.4 & -- \\ \bottomrule
\end{tabular}}
\caption{Statistics of datasets and associated grammars} 
\label{tab:dataset_stat}
\vspace{-3mm}
\end{table}

\noindent \textsc{\textbf{HearthStone}} ({\sc HS}) dataset~\cite{DBLP:conf/acl/LingBGHKWS16} is a collection of Python classes that implement cards for the card game HearthStone. 
Each card comes with a set of fields (e.g., name, cost, and description), which we concatenate to create the input sequence.
This dataset is relatively difficult: input descriptions are short, while the target code is in complex class structures, with each AST having 137 nodes on average.

\noindent \textsc{\textbf{Django}} dataset~\cite{DBLP:conf/kbse/OdaFNHSTN15} is a collection of lines of code from the Django web framework, each with a manually annotated NL description.
Compared with the \hs/ dataset where card implementations are somewhat homogenous, examples in \django/ are more diverse, spanning a wide variety of real-world use cases like string manipulation, IO operations, and exception handling.

\noindent \textsc{\textbf{Ifttt}} dataset~\cite{DBLP:conf/acl/QuirkMG15} is a domain-specific benchmark that provides an interesting side comparison. 
Different from \hs/ and \django/ which are in a general-purpose PL, 
programs in \ifttt/ are written in a domain-specific language used by the IFTTT task automation App.
Users of the App write simple instructions (e.g., {\tt If Instagram.AnyNewPhotoByYou Then Dropbox.AddFileFromURL}) with NL descriptions (e.g., ``{\it Autosave your Instagram photos to Dropbox}'').
Each statement inside the {\tt If} or {\tt Then} clause consists of a channel (e.g., {\tt Dropbox}) and a function (e.g., {\tt AddFileFromURL})\footnote{Like \newcite{DBLP:conf/acl/BeltagyQ16}, we strip function parameters since they are mostly specific to users.}.
This simple structure results in much more concise ASTs (7 nodes on average).
Because all examples are created by ordinary Apps users, the dataset is highly noisy, with input NL very loosely connected to target ASTs.
The authors thus provide a high-quality filtered test set, where each example is verified by at least three annotators. We use this set for evaluation.
Also note \ifttt/'s grammar has more productions (Tab.~\ref{tab:dataset_stat}),
but this does not imply that its grammar is more complex.
This is because for \hs/ and \django/ terminal tokens are generated by {\sc GenToken} actions, but for \ifttt/, all the code is generated directly by {\sc ApplyRule} actions.


\noindent {\bf Metrics} As is standard in semantic parsing, we measure {\bf accuracy}, the fraction of correctly generated examples.
However, because generating an exact match for complex code structures is non-trivial, 
we follow~\newcite{DBLP:conf/acl/LingBGHKWS16}, and use token-level {\bf BLEU-4} with as a secondary metric, defined as the averaged BLEU scores over all examples.%
\footnote{These two metrics are not ideal: accuracy only measures exact match and thus lacks the ability to give credit to semantically correct code that is different from the reference, while it is not clear whether BLEU provides an appropriate proxy for measuring semantics in the code generation task. A more intriguing metric would be directly measuring semantic/functional code equivalence, for which we present a pilot study at the end of this section (cf.~Error Analysis). We leave exploring more sophisticated metrics (e.g. based on static code analysis) as future work.}

\vspace{-1mm}
\subsection{Setup}
\vspace{-1mm}

\noindent {\bf Preprocessing} All input descriptions are tokenized using {\sc nltk}. 
We perform simple canonicalization for \django/, such as replacing quoted strings in the inputs with place holders. See supplementary materials for details.
We extract unary closures whose frequency is larger than a threshold $k$ ($k=30$ for \hs/ and $50$ for \django/).

\noindent {\bf Configuration} The size of all embeddings is 128, except for node type embeddings, which is 64. 
The dimensions of RNN states and hidden layers are 256 and 50, respectively.
Since our datasets are relatively small for a data-hungry neural model, 
we impose strong regularization using recurrent dropouts~\cite{DBLP:conf/nips/GalG16}, 
together with standard dropout layers added to the inputs and outputs of the decoder RNN.
We validate the dropout probability from $\{ 0, 0.2, 0.3, 0.4 \}$.
For decoding, we use a beam size of 15.

\vspace{-1.4mm}
\subsection{Results}
\vspace{-1mm}

Evaluation results for Python code generation tasks are listed in Tab.~\ref{tab:main_results}. 
Numbers for our systems are averaged over three runs.
We compare primarily with two approaches: 
(1) Latent Predictor Network ({\sc lpn}), a state-of-the-art sequence-to-sequence code generation model~\cite{DBLP:conf/acl/LingBGHKWS16}, and
(2) {\sc Seq2Tree}, a neural semantic parsing model~\cite{DBLP:conf/acl/DongL16}. \sq/ generates trees one node at a time, and the target grammar is not explicitly modeled a priori, but {\it implicitly} learned from data.
We test both the original \sq/ model released by the authors and our revised one (\sq/--UNK) that uses unknown word replacement to handle rare words~\cite{DBLP:conf/acl/LuongSLVZ15}.
For completeness, we also compare with a strong neural machine translation ({\sc nmt}) system~\cite{neubig15lamtram} using a standard encoder-decoder architecture with attention and unknown word replacement\footnote{For {\sc nmt}, we also attempted to find the best-scoring syntactically correct predictions in the size-5 beam, but this did not yield a significant improvement over the {\sc nmt} results in Tab.~\ref{tab:main_results}.}, and include numbers from other baselines used in~\newcite{DBLP:conf/acl/LingBGHKWS16}. 
On the \hs/ dataset, which has relatively large ASTs, we use unary closure for our model and \sq/, and for \django/ we do not. 

\begin{table}[]
\centering
\small
\renewcommand{\tabcolsep}{5pt}
\begin{tabular}{@{}lrrrr@{}}
\toprule
                  & \multicolumn{2}{c}{\hs/} & \multicolumn{2}{c}{\django/} \\ \midrule
                  & {\sc acc} & {\sc bleu} & {\sc acc}   & {\sc bleu}   \\
Retrieval System$^\dagger$         & 0.0 & 62.5 & 14.7 & 18.6 \\
Phrasal Statistical MT$^\dagger$            & 0.0 & 34.1 & 31.5 & 47.6 \\
Hierarchical Statistical MT$^\dagger$      & 0.0 & 43.2 & 9.5  & 35.9 \\ \midrule
{\sc nmt}                  & 1.5 & 60.4 & 45.1 & 63.4 \\
\sq/                       & 1.5 & 53.4 & 28.9 & 44.6 \\ 
\sq/--UNK                  & 13.6 & 62.8 & 39.4 & 58.2 \\ 
\lpn/$^\dagger$            & 4.5 & 65.6 & 62.3 & 77.6 \\
\toprule
Our system & 16.2 & {\bf 75.8} & {\bf 71.6} & {\bf 84.5} \\ \midrule
Ablation Study & & & & \\
-- frontier embed. & {\bf 16.7} & {\bf 75.8} & 70.7 & 83.8 \\
-- parent feed.    & 10.6 & 75.7 & 71.5 & 84.3 \\ 
-- copy terminals  & 3.0  & 65.7 & 32.3 & 61.7 \\ 
+ unary closure   & \multicolumn{2}{c}{--} & 70.3 & 83.3 \\
-- unary closure   & 10.1 & 74.8 & \multicolumn{2}{c}{--} \\ \bottomrule
\end{tabular}
\caption{Results on two Python code generation tasks. $^\dagger$Results previously reported in~\newcite{DBLP:conf/acl/LingBGHKWS16}.}
\vspace{-4mm}
\label{tab:main_results}
\end{table}

\noindent {\bf System Comparison}
As in Tab.~\ref{tab:main_results}, our model registers 11.7\% and 9.3\% absolute improvements over \lpn/ in accuracy on \hs/ and \django/.
This boost in performance strongly indicates the importance of modeling grammar in code generation.
For the baselines, we find \lpn/ outperforms others in most cases. 
We also note that \sq/ achieves a decent accuracy of 13.6\% on \hs/, which is due to the effect of unknown word replacement, since we only achieved 1.5\% without it.
A closer comparison with \sq/ is insightful for understanding the advantage of our syntax-driven approach, since both \sq/ and our system output ASTs:
(1) \sq/ predicts one node each time step, and requires additional ``dummy'' nodes to mark the boundary of a subtree. 
The sheer number of nodes in target ASTs makes the prediction process error-prone.
In contrast, the {\sc ApplyRule} actions of our grammar model allows for generating multiple nodes at a single time step.
Empirically, we found that in \hs/, \sq/ takes more than 300 time steps on average to generate a target AST, while our model takes only 170 steps.
(2) 
\sq/ does not directly use productions in the grammar, which possibly leads to grammatically incorrect ASTs and thus empty code outputs. 
We observe that the ratio of grammatically incorrect ASTs predicted by \sq/ on \hs/ and \django/ are 21.2\% and 10.9\%, respectively, while our system guarantees grammaticality.

\begin{figure}[t]
  \centering
  \includegraphics[width=0.85 \columnwidth]{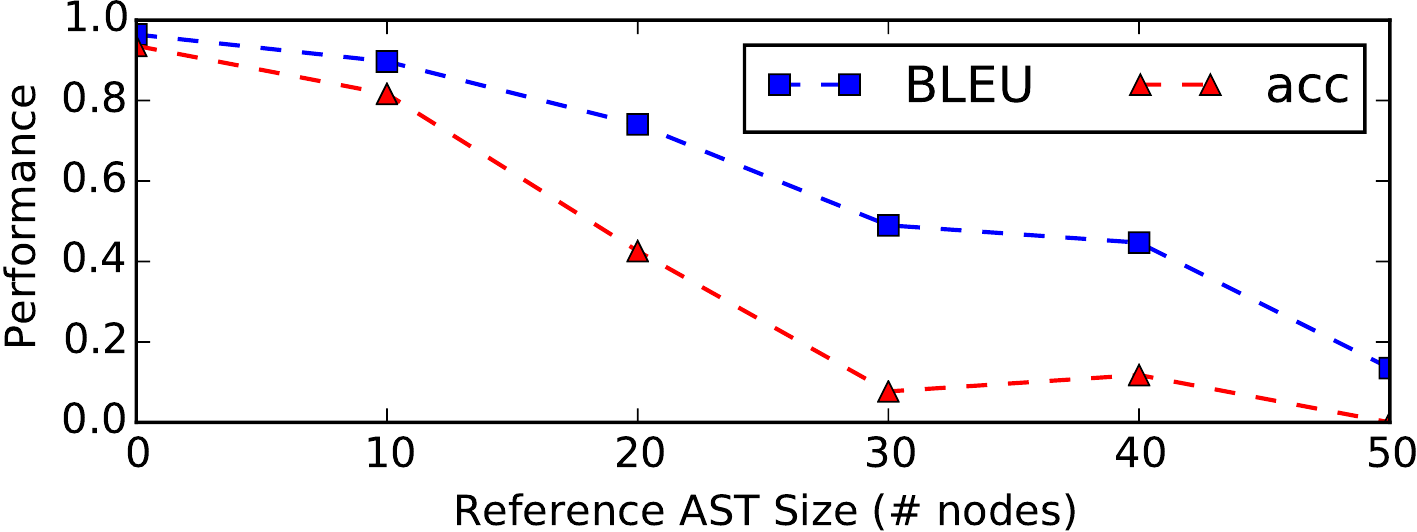}
  \vspace{-2mm}
  \caption{Performance w.r.t reference AST size on \django/}
  \label{fig:django_acc_ast_size}
  \vspace{2mm}
  \includegraphics[width=0.85 \columnwidth]{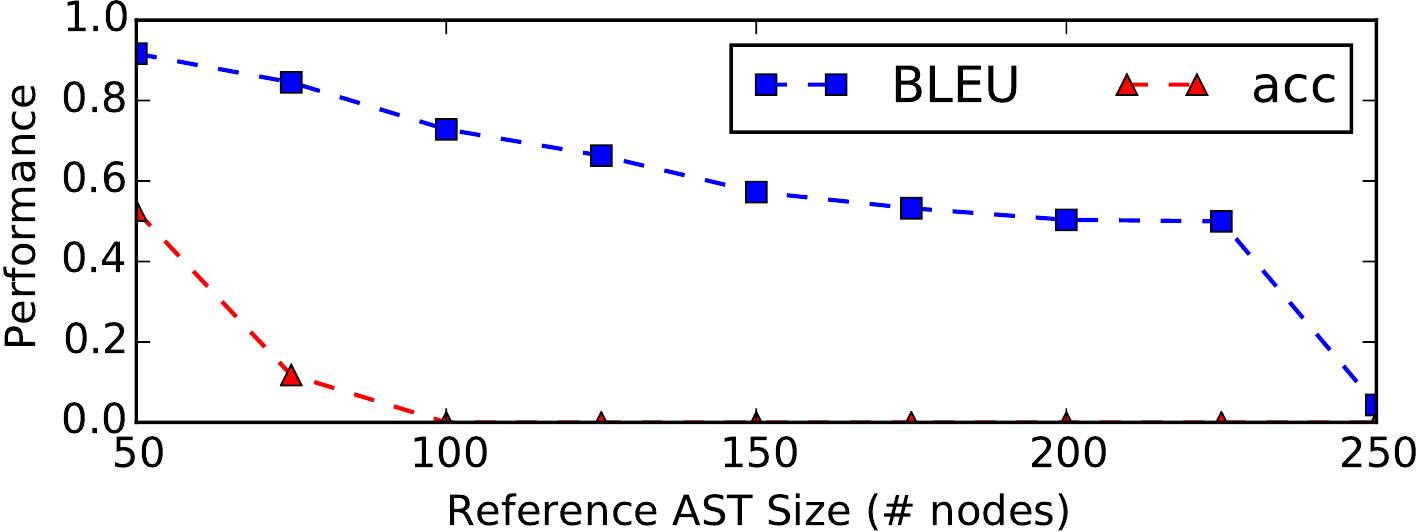}
  \vspace{-2mm}
  \caption{Performance w.r.t reference AST size on \hs/}
  \label{fig:hs_bleu_ast_size}
  \vspace{-4mm}
\end{figure}

\noindent {\bf Ablation Study} We also ablated our best-performing models to analyze the contribution of each component.
``--frontier embed.''~removes the frontier node embedding $\mathbf{n}_{f_t}$ from the decoder RNN inputs (Eq.~\eqref{eq:decoder_lstm}). This yields worse results on \django/ while gives slight improvements in accuracy on \hs/.
This is probably because that the grammar of \hs/ has fewer node types, 
and thus the RNN is able to keep track of $n_{f_t}$ without depending on its embedding.
Next, ``--parent feed.'' removes the parent feeding mechanism.
The accuracy drops significantly on \hs/, with a marginal deterioration on \django/.
This result is interesting because it suggests that parent feeding is more important when the ASTs are larger, which will be the case when handling more complicated code generation tasks like \hs/. 
Finally, removing the pointer network (``--copy terminals'') in {\sc GenToken} actions gives poor results, indicating that it is important to directly copy variable names and values from the input.

The results with and without unary closure demonstrate that, interestingly, it is effective on \hs/ but not on \django/.
We conjecture that this is because on \hs/ it significantly reduces the number of actions from 173 to 142 (c.f., Tab.~\ref{tab:dataset_stat}), with the number of productions in the grammar remaining unchanged.
In contrast, \django/ has a broader domain, and thus unary closure results in more productions in the grammar (237 for \django/ vs.~100 for \hs/), increasing sparsity.

\noindent {\bf Performance by the size of AST} 
We further investigate our model's performance w.r.t.~the size of the gold-standard ASTs in Figs.~\ref{fig:django_acc_ast_size} and~\ref{fig:hs_bleu_ast_size}.
Not surprisingly, the performance drops when the size of the reference ASTs increases.
Additionally, on the \hs/ dataset, the BLEU score still remains at around 50 even when the size of ASTs grows to 200, indicating that our proposed syntax-driven approach is robust for long code segments.

\begin{table}[t]
\centering
\small
\renewcommand{\tabcolsep}{2pt}
\begin{tabular}{@{}lrr@{}}
\toprule
               & {\sc Channel} & {\sc Full Tree} \\ \midrule
{\bf Classical Methods} & \\
posclass~\cite{DBLP:conf/acl/QuirkMG15}       & 81.4 & 71.0 \\ 
LR~\cite{DBLP:conf/acl/BeltagyQ16}            & 88.8 & {\bf 82.5} \\ \midrule

{\bf Neural Network Methods} & \\
{\sc nmt}      & 87.7 & 77.7 \\
NN~\cite{DBLP:conf/acl/BeltagyQ16}            & 88.0 & 74.3 \\
\sq/~\cite{DBLP:conf/acl/DongL16}           & 89.7 & 78.4 \\ 
Doubly-Recurrent NN     & \multirow{ 2}{*}{\bf 90.1} & \multirow{ 2}{*}{78.2} \\
\cite{doublyrnn} \\
\midrule
Our system      & 90.0 & 82.0 \\
-- parent feed. & 89.9 & 81.1 \\
-- frontier embed. & {\bf 90.1} & 78.7 \\ \bottomrule
\end{tabular}
\caption{Results on the noise-filtered \ifttt/ test set of ``$>$3 agree with gold annotations'' (averaged over three runs), our model performs competitively among neural models.}
\label{tab:ifttt_results}
\vspace{-4mm}
\end{table}

\noindent {\bf Domain Specific Code Generation}
Although this is not the focus of our work, evaluation on \ifttt/ brings us closer to a standard semantic parsing setting, which helps to investigate similarities and differences between generation of more complicated general-purpose code and and more limited-domain simpler code.
Tab.~\ref{tab:ifttt_results} shows the results, following the evaluation protocol in~\cite{DBLP:conf/acl/BeltagyQ16} for accuracies at both channel and full parse tree (channel + function) levels.
Our full model performs on par with existing neural network-based methods, while outperforming other neural models in full tree accuracy (82.0\%).
This score is close to the best classical method (LR), which is based on a logistic regression model with rich hand-engineered features (e.g., brown clusters and paraphrase).
Also note that the performance between {\sc nmt} and other neural models is much closer compared with the results in Tab.~\ref{tab:main_results}.
This suggests that general-purpose code generation is more challenging than the simpler \ifttt/ setting, and therefore modeling structural information is more helpful.


\begin{table}[tb]
  \centering
  \small
  \renewcommand{\tabcolsep}{2pt}
  \begin{tabular}{lp{6.7cm}}
  \toprule
  {\bf input} & {\it {\tt <name>} Brawl {\tt </name>} {\tt <cost>} 5 {\tt </cost>} {\tt <desc>} Destroy all minions except one (chosen randomly) {\tt </desc>} {\tt <rarity>} Epic {\tt </rarity>} ... } \\
  {\bf pred.} &
\begin{lstlisting}[basicstyle=\fontfamily{cmtt}\scriptsize,style=pythoncode,belowskip=-\baselineskip,aboveskip=- 0.5\baselineskip]
class |\copyh{Brawl}|(SpellCard):
  def __init__(self):
    super().__init__('|\copyh{Brawl}|', |\copyh{5}|, CHARACTER_CLASS.WARRIOR, CARD_RARITY.EPIC)
  def use(self, player, game):
    super().use(player, game)
    |\tikzmark{starta}|targets = copy.copy(game.other_player.minions)|\tikzmark{enda}|
    targets.extend(player.minions)
    for minion in targets:  
      minion.die(self)
\end{lstlisting}
\begin{tikzpicture}[remember picture,overlay]
\draw[red,rounded corners,dashed]
  ([shift={(-2pt,1.5ex)}]pic cs:starta) 
    rectangle 
  ([shift={(2pt,-6.3ex)}]pic cs:enda);
\draw (pic cs:enda) node[below left=3ex and 2mm] {A};
\end{tikzpicture}
\\
{\bf ref.} & \begin{lstlisting}[basicstyle=\fontfamily{cmtt}\scriptsize,style=pythoncode,belowskip=-\baselineskip,aboveskip=- 0.5\baselineskip]
|\tikzmark{startb}|minions = copy.copy(player.minions)
minions.extend(game.other_player.minions)
if len(minions) > 1:
  survivor = game.random_choice(minions)
  for minion in minions:
    if minion is not survivor: minion.die(self)|\tikzmark{endb}|
\end{lstlisting}
\begin{tikzpicture}[remember picture,overlay]
\draw[red,rounded corners,dashed]
  ([shift={(-2pt,1.5ex)}]pic cs:startb) 
    rectangle 
  ([shift={(2pt,-.6ex)}]pic cs:endb);
\draw (pic cs:endb) node[above left=12mm and 1mm] {B};
\end{tikzpicture}
 \\ \midrule
  {\bf input} & {\it join app\_config.path and string 'locale' into a file path, substitute it for localedir.} \\
  {\bf pred.} & \begin{lstlisting}[basicstyle=\fontfamily{cmtt}\small,style=pythoncode,belowskip=-\baselineskip,aboveskip=- 0.5\baselineskip]
|\copyh{localedir}| = os.path.join(|\copyh{app\_config.path}|, '|\copyh{locale}|')  |\cmark|
  \end{lstlisting} \\
  \midrule
  {\bf input} & {\it self.plural is an lambda function with an argument n, which returns result of boolean expression n not equal to integer 1} \\
  {\bf pred.} & \begin{lstlisting}[basicstyle=\fontfamily{cmtt}\small,style=pythoncode,belowskip=-\baselineskip,aboveskip=- 0.5\baselineskip]
|\copyh{self.plural}| = lambda |\copyh{n}|: len(|\copyh{n}|)  |\xmark|
  \end{lstlisting} \\
  {\bf ref.} & \begin{lstlisting}[basicstyle=\fontfamily{cmtt}\small,style=pythoncode,belowskip=-\baselineskip,aboveskip=- 0.5\baselineskip]
self.plural = lambda n: int(n!=1)
  \end{lstlisting} \\
  \bottomrule
  \end{tabular}
  \caption{Predicted examples from \hs/ (1st) and \django/. Copied contents (copy probability $>0.9$) are highlighted.}
  \label{tab:decode_example}
  \vspace{-4mm}
\end{table}

\noindent {\bf Case Studies} We present output examples in Tab.~\ref{tab:decode_example}.
On \hs/, we observe that most of the time our model gives correct predictions by filling learned code templates from training data with arguments (e.g., cost) copied from input.
However, we do find interesting examples indicating that the model learns to generalize beyond trivial copying.
For instance, the first example is one that our model predicted wrong --- it generated code block A instead of the gold B (it also missed a function definition not shown here). 
However, we find that the block A actually conveys part of the input intent by destroying all, not some, of the minions.
Since we are unable to find code block A in the training data,
it is clear that the model has learned to generalize to some extent from multiple training card examples with similar semantics or structure.

The next two examples are from \django/.
The first one shows that the model learns the usage of common API calls (e.g., {\tt os.path.join}), and how to populate the arguments by copying from inputs.
The second example illustrates the difficulty of generating code with complex nested structures like lambda functions, a scenario worth further investigation in future studies.
More examples are attached in supplementary materials.

\noindent {\bf Error Analysis} To understand the sources of errors and how good our evaluation metric (exact match) is, we randomly sampled and labeled 100 and 50 failed examples (with accuracy=0) from \django/ and \hs/, resp. We found that around 2\% of these examples in the two datasets are actually semantically equivalent. These examples include: (1) using different parameter names when defining a function; (2) omitting (or adding) default values of parameters in function calls. While the rarity of such examples suggests that our exact match metric is reasonable, more advanced evaluation metrics based on statistical code analysis are definitely intriguing future work.

For \django/, we found that 30\% of failed cases were due to errors where the pointer network failed to appropriately copy a variable name into the correct position. 25\% were because the generated code only partially implementated the required functionality. 10\% and 5\% of errors were due to malformed English inputs and pre-processing errors, respectively. The remaining 30\% of examples were errors stemming from multiple sources, or errors that could not be easily categorized into the above. For \hs/, we found that all failed card examples were due to partial implementation errors, such as the one shown in Table~\ref{tab:decode_example}.

\section{Related Work}

\noindent {\bf Code Generation and Analysis}
Most existing works on code generation focus on generating code for domain specific languages (DSLs)~\cite{DBLP:conf/naacl/KushmanB13,DBLP:conf/ijcai/RazaGM15,DBLP:conf/aaai/ManshadiGA13}, with neural network-based approaches recently explored~\cite{DBLP:journals/corr/ParisottoMSLZK16,DBLP:journals/corr/BalogGBNT16}.
For general-purpose code generation, besides the general framework of~\newcite{DBLP:conf/acl/LingBGHKWS16}, existing methods often use language and task-specific rules and strategies~\cite{DBLP:conf/acl/LeiLBR13,DBLP:conf/icse/RaghothamanWH16}.
A similar line is to use NL queries for code retrieval~\cite{building-bing-developer-assistant,DBLP:conf/icml/AllamanisTGW15}. 
The reverse task of generating NL summaries from source code has also been explored~\cite{DBLP:conf/kbse/OdaFNHSTN15,DBLP:conf/acl/IyerKCZ16}.
Finally, there are probabilistic models of source code~\cite{DBLP:conf/icml/MaddisonT14,DBLP:conf/sigsoft/NguyenNNN13}. 
The most relevant work is \newcite{DBLP:conf/icml/AllamanisTGW15}, which uses a factorized model to measure semantic relatedness between NL and ASTs for code retrieval, while our model tackles the more challenging generation task.

\noindent {\bf Semantic Parsing}
Our work is related to the general topic of semantic parsing,
where the target logical forms can be viewed as DSLs.
The parsing process is often guided by grammatical formalisms like combinatory categorical grammars~\cite{DBLP:conf/emnlp/KwiatkowskiCAZ13,DBLP:conf/emnlp/ArtziLZ15}, dependency-based syntax~\cite{DBLP:conf/acl/LiangJK11,pasupat2015compositional} or task-specific formalisms~\cite{DBLP:conf/conll/ClarkeGCR10,DBLP:conf/acl/YihCHG15,DBLP:conf/emnlp/KrishnamurthyTK16,DBLP:conf/acl/MisraTLS15,DBLP:conf/aaai/MeiBW16}. 
Recently, there are efforts in designing neural network-based semantic parsers~\cite{Misra:16neuralccg,DBLP:conf/acl/DongL16,DBLP:journals/corr/NeelakantanLS15,DBLP:dblp_conf/ijcai/YinLLK16}.
Several approaches have be proposed to utilize grammar knowledge in a neural parser, such as augmenting the training data by generating examples guided by the grammar~\cite{DBLP:conf/emnlp/KociskyMGDLBH16,DBLP:conf/acl/JiaL16}.
\newcite{DBLP:journals/corr/LiangBLFL16} used a neural decoder which constrains the space of next valid tokens in the query language for question answering.
Finally, the structured prediction approach proposed by \newcite{DBLP:conf/acl/XiaoDG16} is closely related to our model in using the underlying grammar as prior knowledge to constrain the generation process of derivation trees, while our method is based on a unified grammar model which jointly captures production rule application and terminal symbol generation, and scales to general purpose code generation tasks. 

\vspace{-2mm}
\section{Conclusion}
\vspace{-2mm}

This paper proposes a syntax-driven neural code generation approach that 
generates an abstract syntax tree by sequentially applying actions from a grammar model.
Experiments on both code generation and semantic parsing tasks demonstrate the effectiveness of our proposed approach. 

\vspace{-2mm}
\section*{Acknowledgment}
\vspace{-2mm}

We are grateful to Wang Ling for his generous help with \lpn/ and setting up the benchmark. We also thank Li Dong for helping with \sq/ and insightful discussions.

\bibliography{code_gen}
\bibliographystyle{acl_natbib}

\clearpage
\newpage
\begin{center}
  \textbf{\Large{Supplementary Materials}}
\end{center}
\appendix
\input{appendix}

\end{document}

%% file: settings.tex
\usepackage{pifont}
\usepackage{url}
\usepackage{tikz}
\usepackage{latexsym}
\usepackage{color}
\usepackage{enumitem}
\usepackage{graphicx}
\usepackage{amsmath,amsfonts,amssymb}
\usepackage{listings}
\usepackage{subfigure}
\usepackage[ruled,noresetcount,linesnumbered]{algorithm2e}
\usepackage{scalerel}
\usepackage{booktabs}
\usepackage{multirow}
\usepackage{pbox}
\usepackage{todonotes}
\usepackage[font=small]{caption}
\usepackage{listings}
\usepackage[export]{adjustbox}
\usetikzlibrary{tikzmark}

\definecolor{darkgreen}{RGB}{43,163,39}
\newcommand{\cmark}{\textcolor{darkgreen}{\ding{51}}}
\newcommand{\xmark}{\textcolor{red}{\ding{55}}}
\DeclareFixedFont{\ttb}{T1}{txtt}{bx}{n}{12} 
\DeclareFixedFont{\ttm}{T1}{txtt}{m}{n}{12}  

\usepackage{color}
\definecolor{deepblue}{rgb}{0,0,0.5}
\definecolor{deepred}{rgb}{0.6,0,0}
\definecolor{deepgreen}{rgb}{0,0.5,0}

\newcommand\mybox[2][]{\tikz[overlay]\node[inner sep=1pt, anchor=text, rectangle, rounded corners=1mm,#1] {#2};\phantom{#2}}
\definecolor{fillcolor}{RGB}{216,217,252}
\newcommand\copyh[1]{\mybox[fill=blue!20]{#1}}

\lstdefinestyle{pythoncode}{
	language=Python,
	otherkeywords={self},             
	keywordstyle=\bfseries\color{deepblue},
	emph={MyClass,__init__},          
	emphstyle=\color{deepred},    
	showstringspaces=false,
	breaklines=true,
	escapeinside=||,
	columns=fullflexible,
}

\newcommand\wrong[1]{\textcolor{red}{#1}}

\lstdefinestyle{django}{
  language=Python,
  otherkeywords={self},             
  keywordstyle=\bfseries\color{deepblue},
  emph={MyClass,__init__},          
  emphstyle=\color{deepred},    
  showstringspaces=false,
  breaklines=true,
  escapeinside=||,
  columns=fullflexible,
  basicstyle=\fontfamily{cmtt}\small,
  belowskip=-\baselineskip,
  aboveskip=-0.7\baselineskip
}

\renewcommand{\tt}[1]{\fontfamily{cmtt}\selectfont #1}
\newcommand{\cev}[1]{\reflectbox{\ensuremath{\vec{\reflectbox{\ensuremath{#1}}}}}}
\DeclareMathOperator*{\argmax}{arg\,max}

\def\lpn/{{\sc lpn}}
\def\sq/{{\sc Seq2Tree}}
\def\django/{\textsc{Django}}
\def\hs/{\textsc{HS}}
\def\ifttt/{\textsc{Ifttt}}

%% file: appendix.tex
\section{Encoder LSTM Equations}
\label{app:encoder}

Suppose the input natural language description $x$ consists of $n$ words $\{w_i\}_{i=1}^n$.
Let $\mathbf{w}_i$ denote the embedding of $w_i$.
We use two LSTMs to process $x$ in forward and backward order, and get the sequence of hidden states $\{\vec{\mathbf{h}}_i\}_{i=1}^{n}$ and $\{\cev{\mathbf{h}}_i\}_{i=1}^{n}$ in the two directions:
\begin{equation*}
\begin{split}
  \vec{\mathbf{h}}_i &= f^{\to}_\textrm{LSTM}(\mathbf{w}_i, \vec{\mathbf{h}}_{i-1}) \\
  \cev{\mathbf{h}}_i &= f^{\gets}_\textrm{LSTM}(\mathbf{w}_{i}, \cev{\mathbf{h}}_{i+1}),
\end{split}
\end{equation*}
where $f^{\to}_\textrm{LSTM}$ and $f^{\gets}_\textrm{LSTM}$ are standard LSTM update functions.
The representation of the $i$-th word, $\mathbf{h}_i$, is given by concatenating $\vec{\mathbf{h}}_i$ and $\cev{\mathbf{h}}_i$.

\section{Inference Algorithm}
\label{app:inference}

Given an NL description, we approximate the best AST $\hat{y}$ in Eq.~1 using beam search. The inference procedure is listed in Algorithm~\ref{algo:inference}.

We maintain a beam of size $K$.
The beam is initialized with one hypothesis AST with a single root node (line~\ref{algo:inference.initQ}).
At each time step, the decoder enumerates over all hypotheses in the beam.
For each hypothesis AST, we first find its frontier node $n_{f_t}$ (line~\ref{algo:inference.fn}).
If $n_{f_t}$ is a non-terminal node, we collect all syntax rules $r$ with $n_{f_t}$ as the head node to the actions set (line~\ref{algo:inference.nonterminal}).
If $n_{f_t}$ is a variable terminal node, we add all terminal tokens in the vocabulary and the input description as candidate actions (line~\ref{algo:inference.terminal}).
We apply each candidate action on the current hypothesis AST to generate a new hypothesis (line~\ref{algo:inference.newHyp}).
We then rank all newly generated hypotheses and keep the top-$K$ scored ones in the beam.
A complete hypothesis AST is generated when it has no frontier node.
We then convert the top-scored complete AST into the surface code (lines~\ref{algo:inference.completeAST}-\ref{algo:inference.tree2code}).

\begin{algorithm*}[h]
  \caption{Inference Algorithm}
  \label{algo:inference}
  \small
  \DontPrintSemicolon
  \SetKwInOut{Input}{Input}
  \SetCommentSty{itshape}
  \SetKwComment{Comment}{$\triangleright$\ }{}
  \SetAlgoNoEnd
  \SetKwInOut{Output}{Output}
  \Input{NL description $x$}
  \Output{code snippet $c$}
  call \textsf{Encoder} to encode $x$\;
  $Q = \{ \mathit{y_0}(\mathit{root}) \}$ \Comment*[r]{Initialize a beam of size $K$} \label{algo:inference.initQ}
  \For {time step $t$}
  {
    $Q' = \emptyset$\;
    \ForEach {hypothesis $y_t \in Q$} { \label{algo:inference.batchbegin}
      $n_{f_t} = \textsf{FrontierNode}(y_t)$ \label{algo:inference.fn}\;
      $\mathcal{A} = \emptyset$ \Comment*[r]{Initialize the set of candidate actions}
      \If {$n_{f_t}$ is non-terminal} { 
        \hspace{-2mm} \ForEach {production rule $r$ with $n_{f_t}$ as the head node} {
          $\mathcal{A} = \mathcal{A} \cup \{ \textsc{ApplyRule}[r] \}$ \label{algo:inference.nonterminal} \Comment*[r]{{\sc ApplyRule} actions for non-terminal nodes}
        }
      }
      \Else {
        \ForEach {terminal token $v$} {
          $\mathcal{A} = \mathcal{A} \cup \{ \textsc{GenToken}[v] \}$ \label{algo:inference.terminal} \Comment*[r]{{\sc GenToken} actions for variable terminal nodes}
        }
      }

      \ForEach {action $a_t \in \mathcal{A}$} {
        $y_t' = \textsf{ApplyAction}(y_t, a_t)$ \label{algo:inference.newHyp} \;
        $Q' = Q' \cup \{ \mathit{y_t'} \}$\;
      }
    } \label{algo:inference.batchend}
    $Q = \textrm{top-$K$ scored hypotheses in $Q'$}$\;
  }
  $\hat{y} = \textrm{top-scored complete hypothesis AST}$ \label{algo:inference.completeAST}\;
  \textrm{convert $\hat{y}$ to surface code $c$} \label{algo:inference.tree2code}\;
  \Return{$c$}
\end{algorithm*}

We remark that our inference algorithm can be implemented efficiently by expanding multiple hypotheses (lines \ref{algo:inference.batchbegin}-\ref{algo:inference.batchend}) simultaneously using mini-batching on GPU. 

\section{Dataset Preprocessing}

\noindent {\bf Infrequent Words} We replace word types whose frequency is lower than $d$ with a special {\tt <unk>} token ($d=3$ for \django/, 3 for \hs/ and 2 for \ifttt/). 

\noindent {\bf Canonicalization} We perform simple canonicalization for the \django/ dataset: (1) We observe that input descriptions often come with quoted string literals (e.g., {\it verbose\_name is a string \underline{`cache entry'}}). We therefore replace quoted strings with indexed placeholders using regular expression. After decoding, we run a post-processing step to replace all placeholders with their actual values. (2) For descriptions with cascading variable reference (e.g., {\it call method \underline{self.makekey}}), we append after the whole variable name with tokens separated by `.' (e.g., append {\it self} and {\it makekey} after {\it self.makekey}). This gives the pointer network flexibility to copy either partial or whole variable names.

\noindent {\bf Generate Oracle Action Sequence} To train our model, we generate the gold-standard action sequence from reference code.
For \ifttt/, we simply parse the officially provided ASTs into sequences of {\sc ApplyRule} actions. 
For \hs/ and \django/, we first convert the Python code into ASTs using the standard {\tt ast} module.
Values inside variable terminal nodes are tokenized by space and camel case (e.g., {\it ClassName} is tokenized to {\it Class} and {\it Name}).
We then traverse the AST in pre-order to generate the reference action sequence according to the grammar model.




\begin{table*}[h]
\centering
\footnotesize
\renewcommand{\tabcolsep}{2pt}
  \footnotesize
  \centering
  \renewcommand{\tabcolsep}{1.6pt}
  \hskip-0.3cm
  \begin{tabular}{rp{7.0cm}rp{7.7cm}}
  \toprule
  {\bf input} & \multicolumn{3}{l}{{\it for every i in range of integers from 0 to length of result, not included}} \\
  {\bf pred.} & 
\begin{lstlisting}[style=django]
for |\copyh{i}| in range(|\copyh{0}|, len(|\copyh{result}|)): |\cmark|
\end{lstlisting} & {\bf ref.} &
\begin{lstlisting}[style=django]
for i in range(len(result)):
\end{lstlisting} \\ \midrule
{\bf input} & \multicolumn{3}{l}{{\it call the function blankout with 2 arguments: t.contents and 'B', write the result to out.}} \\
  {\bf pred.} & 
\begin{lstlisting}[style=django]
|\copyh{out}|.|\copyh{write}|(|\copyh{blankout}|(|\copyh{t.contents}|, '|\copyh{B}|')) |\cmark|
\end{lstlisting} & {\bf ref.} &
\begin{lstlisting}[style=django]
out.write(blankout(t.contents, 'B'))
\end{lstlisting} \\ \midrule
  {\bf pred.} & 
\begin{lstlisting}[style=django]
|\copyh{code\_list}|.|\copyh{append}|(|\copyh{foreground}|[|\copyh{v}|]) |\cmark|
\end{lstlisting} & {\bf ref.} &
\begin{lstlisting}[style=django]
code_list.append(foreground[v])
\end{lstlisting} \\ \midrule
{\bf input} & \multicolumn{3}{l}{{\it zip elements of inner\_result and inner\_args into a list of tuples, for every i\_item and i\_args in the result}} \\
  {\bf pred.} & 
\begin{lstlisting}[style=django]
for |\copyh{i\_item}|, |\copyh{i\_args}| in |\copyh{zip}|(|\copyh{inner\_result}|, |\copyh{inner\_args}|): |\cmark|
\end{lstlisting} & {\bf ref.} &
\begin{lstlisting}[style=django]
for i_item, i_args in zip(inner_result, inner_args):
\end{lstlisting} \\ \midrule
{\bf input} & \multicolumn{3}{l}{{\it activate is a lambda function which returns None for any argument x.}} \\
  {\bf pred.} & 
\begin{lstlisting}[style=django]
|\copyh{activate}| = lambda |\copyh{x}|: |\copyh{None}| |\cmark|
\end{lstlisting} & {\bf ref.} &
\begin{lstlisting}[style=django]
activate = lambda x: None
\end{lstlisting} \\ \midrule
{\bf input} & \multicolumn{3}{l}{{\it if elt is an instance of Choice or NonCapture classes}} \\
  {\bf pred.} & 
\begin{lstlisting}[style=django]
if isinstance(|\copyh{elt}|, |\copyh{Choice}|): |\xmark|
\end{lstlisting} & {\bf ref.} &
\begin{lstlisting}[style=django]
if isinstance(elt, (Choice, NonCapture)):
\end{lstlisting} \\ \midrule
  {\bf input} & \multicolumn{3}{p{15cm}}{{\it get translation\_function attribute of the object t, call the result with an argument eol\_message, substitute the result for result.}} \\
  {\bf pred.} & 
\begin{lstlisting}[style=django]
|\copyh{translation\_function}| = getattr(|\copyh{t}|, |\copyh{translation\_function}|) |\xmark|
\end{lstlisting} & {\bf ref.} &
\begin{lstlisting}[style=django]
result = getattr(t, translation_function)(eol_message)
\end{lstlisting} \\ \midrule
{\bf input} & \multicolumn{3}{l}{{\it for every s in strings, call the function force\_text with an argument s, join the results in a string, return the result.}} \\
  {\bf pred.} & 
\begin{lstlisting}[style=django]
return ''.join(|\copyh{force\_text}|(|\copyh{s}|)) |\xmark|
\end{lstlisting} & {\bf ref.} &
\begin{lstlisting}[style=django]
return ''.join(force_text(s) for s in strings)
\end{lstlisting} \\ \midrule
{\bf input} & \multicolumn{3}{l}{{\it for every p in parts without the first element}} \\
  {\bf pred.} & 
\begin{lstlisting}[style=django]
for |\copyh{p}| in |\copyh{p}|[1:]: |\xmark|
\end{lstlisting} & {\bf ref.} &
\begin{lstlisting}[style=django]
for p in parts[1:]:
\end{lstlisting} \\ \midrule
{\bf input} & \multicolumn{3}{l}{{\it call the function get\_language, split the result by '-', substitute the first element of the result for base\_lang.}} \\
  {\bf pred.} & 
\begin{lstlisting}[style=django]
|\copyh{base\_lang}| = |\copyh{get\_language}|().|\copyh{split}|()[0] |\xmark|
\end{lstlisting} & {\bf ref.} &
\begin{lstlisting}[style=django]
base_lang = get_language().split('-')[0]
\end{lstlisting} \\
  \bottomrule
  \end{tabular}

  \caption{Predicted examples from \django/ dataset. Copied contents (copy probability $>0.9$) are highlighted}
  \label{tab:decode_example_django}
\end{table*}

\section{Additional Decoding Examples}

We provide extra decoding examples from the \django/ and \hs/ datasets, listed in Table~\ref{tab:decode_example_django} and Table~\ref{tab:decode_example_hs}, respectively.
The model heavily relies on the pointer network to copy variable names and constants from input descriptions.
We find the source of errors in \django/ is more diverse, with most incorrect examples resulting from missing arguments and incorrect words copied by the pointer network.
Errors in \hs/ are mostly due to partially or incorrectly implemented effects. 
Also note that the first example in Table~\ref{tab:decode_example_django} is semantically correct, although it was considered incorrect under our exact-match metric. This suggests more advanced evaluation metric that takes into account the execution results in future studies.

\begin{table*}[tb]
  \centering
  \footnotesize
  \renewcommand{\tabcolsep}{2pt}
  \begin{tabular}{rp{15cm}}
  \toprule
  {\bf input} & {\it {\tt <name>} Burly Rockjaw Trogg {\tt </name>} {\tt <cost>} 5 {\tt </cost>} {\tt <attack>} 3 {\tt </attack>} {\tt <defense>} 5 {\tt </defense>} {\tt <desc>} Whenever your opponent casts a spell, gain 2 Attack. {\tt </desc>} {\tt <rarity>} Common {\tt </rarity>} ... } \\
  {\bf pred.} &
\begin{lstlisting}[basicstyle=\fontfamily{cmtt}\scriptsize,style=pythoncode,belowskip=-\baselineskip,aboveskip=- 0.5\baselineskip]
class |\copyh{BurlyRockjawTrogg}|(MinionCard):
    def __init__(self):
        super().__init__('|\copyh{Burly}| |\copyh{Rockjaw}| |\copyh{Trogg}|', |\copyh{4}|, CHARACTER_CLASS.ALL, CARD_RARITY.COMMON)
    def create_minion(self, player):
        return Minion(|\copyh{3}|, |\copyh{5}|, effects=[Effect(SpellCast(player=EnemyPlayer()),
            ActionTag(Give(ChangeAttack(|\copyh{2}|)), SelfSelector()))])    |\cmark|
\end{lstlisting} \\ \midrule
  {\bf input} & {\it {\tt <name>} Maexxna {\tt </name>} {\tt <cost>} 6 {\tt </cost>} {\tt <attack>} 2 {\tt </attack>} {\tt <defense>} 8 {\tt </defense>} {\tt <desc>} Destroy any minion damaged by this minion. {\tt </desc>} {\tt <rarity>} Legendary {\tt </rarity>} ... } \\
  {\bf pred.} &
\begin{lstlisting}[basicstyle=\fontfamily{cmtt}\scriptsize,style=pythoncode,belowskip=-\baselineskip,aboveskip=- 0.5\baselineskip]
class |\copyh{Maexxna}|(MinionCard):
    def __init__(self):
        super().__init__('|\copyh{Maexxna}|', |\copyh{6}|, CHARACTER_CLASS.ALL, CARD_RARITY.LEGENDARY, 
          minion_type=MINION_TYPE.BEAST)
    def create_minion(self, player):
        return Minion(|\copyh{2}|, |\copyh{8}|, effects=[Effect(DidDamage(), ActionTag(Kill(),
            TargetSelector(IsMinion())))])   |\cmark|
\end{lstlisting} \\ \midrule \midrule
  {\bf input} & {\it {\tt <name>} Hellfire {\tt </name>} {\tt <cost>} 4 {\tt </cost>} {\tt <attack>} -1 {\tt </attack>} {\tt <defense>} -1 {\tt </defense>} {\tt <desc>} Deal 3 damage to ALL characters. {\tt </desc>} {\tt <rarity>} Free {\tt </rarity>} ... } \\
  {\bf pred.} &
\begin{lstlisting}[basicstyle=\fontfamily{cmtt}\scriptsize,style=pythoncode,belowskip=-\baselineskip,aboveskip=- 0.5\baselineskip]
class |\copyh{Hellfire}|(SpellCard):
    def __init__(self):
        super().__init__('|\copyh{Hellfire}|', |\copyh{4}|, CHARACTER_CLASS.WARLOCK, CARD_RARITY.FREE)

    def use(self, player, game):
        super().use(player, game)
        |\wrong{for minion in copy.copy(game.other\_player.minions):}|
            minion.damage(player.effective_spell_damage(|\copyh{3}|), self)  |\xmark|
\end{lstlisting} \\ 
  {\bf ref.} &
\begin{lstlisting}[basicstyle=\fontfamily{cmtt}\scriptsize,style=pythoncode,belowskip=-\baselineskip,aboveskip=- 0.5\baselineskip]
class Hellfire(SpellCard):
    def __init__(self):
        super().__init__('Hellfire', 4, CHARACTER_CLASS.WARLOCK, CARD_RARITY.FREE)

    def use(self, player, game):
        super().use(player, game)
        targets = copy.copy(game.other_player.minions)
        targets.extend(game.current_player.minions)
        targets.append(game.other_player.hero)
        targets.append(game.current_player.hero)
        for minion in targets:
            minion.damage(player.effective_spell_damage(3), self)
\end{lstlisting} \\
  {\bf reason} & Partially implemented effect: only deal 3 damage to opponent's characters  \\ \midrule
  {\bf input} & {\it {\tt <name>} Darkscale Healer {\tt </name>} {\tt <cost>} 5 {\tt </cost>} {\tt <attack>} 4 {\tt </attack>} {\tt <defense>} 5 {\tt </defense>} {\tt <desc>} Battlecry: Restore 2 Health to all friendly characters. {\tt </desc>} {\tt <rarity>} Common {\tt </rarity>} ... } \\
  {\bf pred.} &
\begin{lstlisting}[basicstyle=\fontfamily{cmtt}\scriptsize,style=pythoncode,belowskip=-\baselineskip,aboveskip=- 0.5\baselineskip]
class |\copyh{DarkscaleHealer}|(MinionCard):
    def __init__(self):
        super().__init__('|\copyh{Darkscale}| |\copyh{Healer}|', |\copyh{5}|, CHARACTER_CLASS.ALL,
            CARD_RARITY.COMMON, battlecry=Battlecry(|\wrong{Damage(\copyh{2})}|,
            |\wrong{CharacterSelector(players=BothPlayer(), picker=UserPicker())}|))

    def create_minion(self, player):
        return Minion(|\copyh{4}|, |\copyh{5}|)  |\xmark|
\end{lstlisting} \\ 
  {\bf ref.} &
\begin{lstlisting}[basicstyle=\fontfamily{cmtt}\scriptsize,style=pythoncode,belowskip=-\baselineskip,aboveskip=- 0.5\baselineskip]
class DarkscaleHealer(MinionCard):
    def __init__(self):
        super().__init__('Darkscale Healer', 5, CHARACTER_CLASS.ALL,
            CARD_RARITY.COMMON, battlecry=Battlecry(Heal(2), CharacterSelector()))

    def create_minion(self, player):
        return Minion(4, 5)
\end{lstlisting} \\
  {\bf reason} & Incorrect effect: damage 2 health instead of restoring. Cast effect to all players instead of friendly players only.  \\
  \bottomrule
  \end{tabular}
  \caption{Predicted card examples from \hs/ dataset. Copied contents (copy probability $>0.9$) are highlighted.}
  \label{tab:decode_example_hs}
\end{table*}